\documentclass[letterpaper]{article} 
\usepackage{aaai24}  
\usepackage{times}  
\usepackage{helvet}  
\usepackage{courier}  
\usepackage[hyphens]{url}  
\usepackage{graphicx} 
\usepackage{natbib}  
\usepackage{caption} 
\usepackage{algorithm}
\usepackage{algorithmic}

\usepackage{xcolor}
\usepackage{multirow}
\usepackage{amsmath}
\usepackage{booktabs}
\usepackage{bm}
\usepackage{mathrsfs}
\usepackage{bbding}
\usepackage{amssymb}
\usepackage{pifont}
\usepackage{wasysym}

\usepackage{colortbl}
\definecolor{mygray}{gray}{.9}
\definecolor{mypink}{rgb}{.99,.91,.95}
\definecolor{mycyan}{cmyk}{.3,0,0,0}
\definecolor{beige}{rgb}{0.96, 0.96, 0.86}
\definecolor{cadmiumgreen}{rgb}{0.0, 0.42, 0.24}
\definecolor{amber}{rgb}{1.0, 0.75, 0.0}
\definecolor{emerald}{rgb}{0.31, 0.78, 0.47}

\newcommand{\ie}{\textit{i}.\textit{e}., }
\newcommand{\eg}{\textit{e}.\textit{g}., }
\newcommand{\etc}{\textit{etc}}

\usepackage[accsupp]{axessibility}

\usepackage{newfloat}
\usepackage{listings}
\DeclareCaptionStyle{ruled}{labelfont=normalfont,labelsep=colon,strut=off} 
\floatstyle{ruled}
\newfloat{listing}{tb}{lst}{}
\floatname{listing}{Listing}
\title{Spatial-Semantic Collaborative Cropping for User Generated Content}
\author{
    Yukun Su\textsuperscript{\rm 1,\rm 2},
    Yiwen Cao\textsuperscript{\rm 1},
    Jingliang Deng\textsuperscript{\rm 1},
    Fengyun Rao\textsuperscript{\rm 2},
    Qingyao Wu\textsuperscript{\rm 1,\rm 3}\thanks{Corresponding author.}
}
\affiliations{
    \textsuperscript{\rm 1} School of Software and Engineering, South China University of Technology\\
    \textsuperscript{\rm 2} WeChat, Tencent Inc.\\
    \textsuperscript{\rm 3} Key Laboratory of Big Data and Intelligent Robot, Ministry of Education\\

    suyukun666@outlook.com; fengyunrao@tencent.com; qyw@scut.edu.cn
}

\usepackage{bibentry}

\begin{document}

\maketitle

\begin{abstract}
A large amount of User Generated Content (UGC) is uploaded to the Internet daily and displayed to people world-widely through the client side (e.g., mobile and PC).
This requires the cropping algorithms to produce the aesthetic thumbnail within a specific aspect ratio on different devices. However, existing image cropping works mainly focus on landmark or landscape images, which fail to model the relations among the multi-objects with the complex background in UGC. Besides, previous methods merely consider the aesthetics of the cropped images while ignoring the content integrity, which is crucial for UGC cropping. In this paper, we propose a Spatial-Semantic Collaborative cropping network (S$^2$CNet) for arbitrary user generated content accompanied by a new cropping benchmark. Specifically, we first mine the visual genes of the potential objects. Then, the suggested adaptive attention graph recasts this task as a procedure of information association over visual nodes. The underlying spatial and semantic relations are ultimately centralized to the crop candidate through differentiable message passing, which helps our network efficiently to preserve both the aesthetics and the content integrity. Extensive experiments on the proposed UGCrop5K and other public datasets demonstrate the superiority of our approach over state-of-the-art counterparts. Our project is available at \url{https://github.com/suyukun666/S2CNet}.

\end{abstract}

\section{Introduction}

\begin{figure}[t]
	\begin{center}
		\centering
		\includegraphics[width=3.0in]{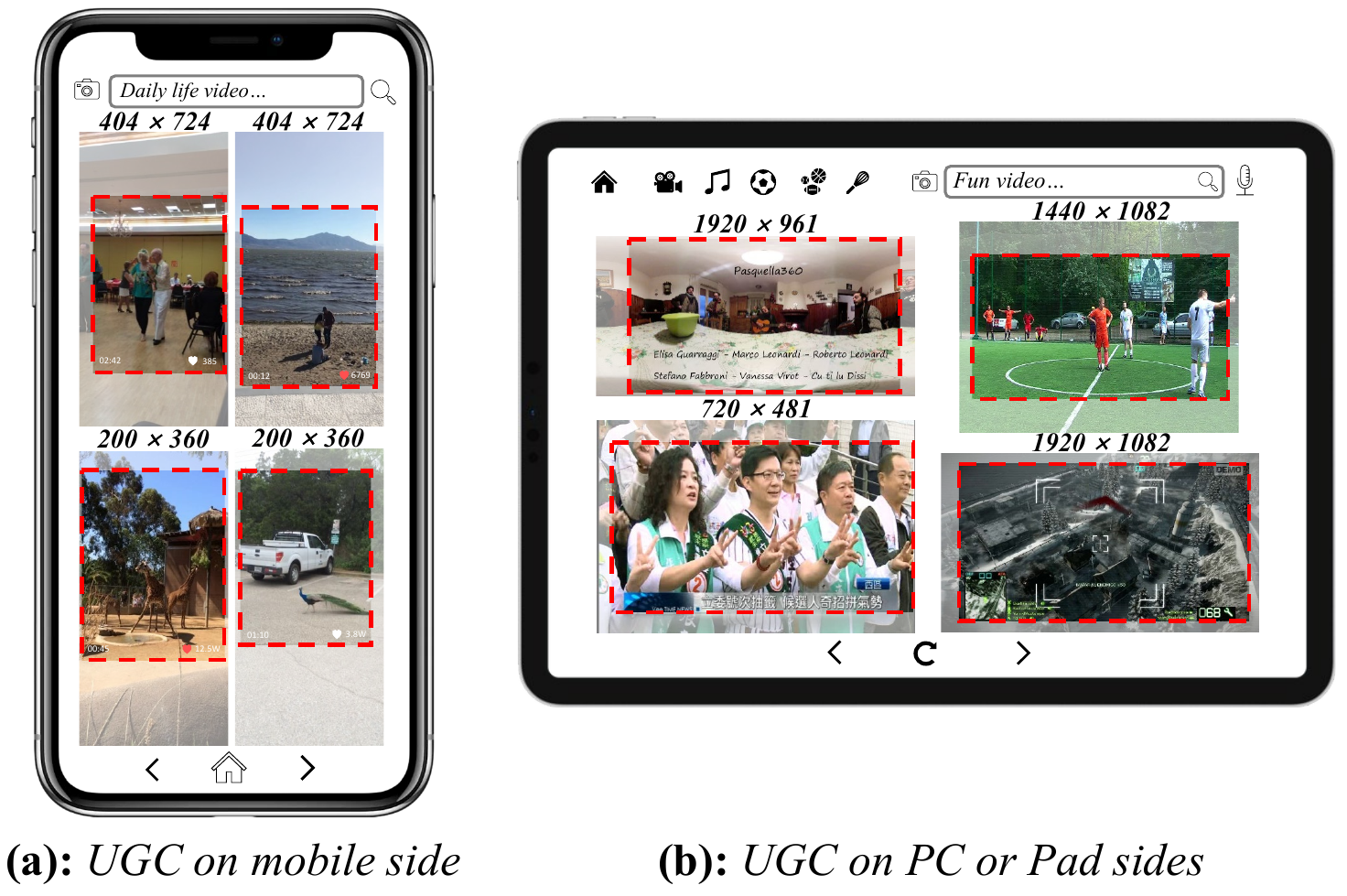}
	\end{center}
	\caption{{{Illustrative example of cropping for UGC in a real-life application, which is more complicated with multi-objects and confounding backgrounds ranging from life clips, news, sports, games and lyric videos, \etc.} Note that the original size of the UGC is marked above each image. For intuitive explanation, the red dashed box indicates the cropped image produced by our algorithm for a fixed aspect ratio and the extraneous content is removed. Best viewed by zooming.}}
	\label{fig1}
\end{figure}

Image cropping, with the aim to automatically excavate appealing views in photography, which integrates some segmentation and detection strategies~\cite{su2021context,su2023unified2} and is widely used for image aesthetic compositions such as thumbnail generation~\cite{chen2018cropnet,esmaeili2017fast}, shot recommendation~\cite{li2018a2,wei2018good} and portrait suggestion~\cite{zhang2018pose,yee2021image}, \etc. Among them, image thumbnailing or cover cropping is a vital application for the explosive emerging User Generated Content (\textbf{\textit{UGC}}).
Since users upload their self-created images or videos to the social media platform using different types of shooting equipment with lenses of various aspect ratios, as shown in Fig~\ref{fig1}, this requires the cropping algorithms to generate the fixed aspect ratios cover images for content aesthetics and format unity.

However, several previous works~\cite{chen2017learning,wei2018good,zeng2019reliable,zeng2020grid,pan2021transview,jia2022rethinking} mainly focus on some iconic landscape images~\cite{chen-wacv2017,zeng2020grid} collected from \textit{Flickr} or even some human-centric~\cite{zhang2022human} images, where these images have clean backgrounds and they are relatively simple to crop. 
By contrast, the main challenges of cropping the user generated content are three folds: (\textbf{{i}}) UGC is more complex with different foreground multi-objects and chaotic backgrounds, thus it's necessary to mine the relations between different objects to find the appealing crops. Meanwhile, some of the saliency-based cropping methods~\cite{chen2016automatic,tu2020image,zhang2022human,cheng2022re,su2023unified} may fail to locate the accurate content; (\textbf{{ii}}) In addition to ensuring the aesthetics of the cropped images, content integrity is also crucial, which conveys the main message to the viewers. As shown in Fig~\ref{fig1}(b), for some news clips or lyric videos, the cropping target should retain the main attributes of the image except for the people, such as the news headline and the complete lyrics. As for the multi-people images, incomplete face cases should be avoided; (\textbf{{iii}}) UGC cropping usually requires the fixed aspect ratio image output for display. Therefore, some anchor-generation-based methods~\cite{hong2021composing,jia2022rethinking} are unsuitable since they follow the process like object detection~\cite{carion2020end} and yield the crop candidates without aspect ratio constraint, which inevitably hinders their application in real-world scenarios.

The common practice of image cropping is to rank candidate views by data-driven methods using
deep neural networks~\cite{simonyan2014very,sandler2018mobilenetv2}. Although numerous efforts have been made, due to the limitations of the public datasets and the technical solutions, many existing methods fail to achieve satisfactory performance for user generated content. 
To learn the relations between different image regions, an alternative solution is to utilize graph convolution networks (GCNs)~\cite{chen2020simple}. Li $\emph{et~al.}$~\cite{li2020composing} exploited relations between different candidates with a graph-based module. However, it does not consider the instance-level cues in the images and the vanilla GCN may lead to the over-smoothing phenomenon. Pan $\emph{et~al.}$~\cite{pan2021transview} adopted the vision transformer (ViT)~\cite{dosovitskiy2020vit} to model the visual element dependencies. Yet, the original ViT ignores the edge and spatial information between the unstructured data. 

To tackle the issues mentioned above, we propose a \underline{S}patial-\underline{S}emantic \underline{C}ollaborative cropping network (\textbf{\textit{S$^2$CNet}}) to effectively crop arbitrary user generated content.
Firstly, we mine the visual genes of the potential objects utilizing the off-the-shelf approach~\cite{ren2015faster} to obtain region-of-interests (RoIs). Afterwards, they are used as the bias combined with the crop candidate and are fed into the proposed framework.
Specifically, we design an adaptive attention graph where each RoI is viewed as the \textit{node} and the correlation between each other is represented as the \textit{edge}. Unlike prior works, we build the graph considering semantic and spatial collaborative information to capture both feature appearance and topological composition representations. Furthermore, we modify the graph convolution operation into a graph-aware attention module to efficiently model the high-order relations among each RoI, which recasts the network as a procedure of information association over visual nodes. The updated messages are ultimately centralized to the crop candidate for aesthetic score prediction. Furthermore, we also construct a large \textbf{\textit{UGCrop5K}} dataset to fill the gap in the image cropping domain, which contains 450,000 exhaustive annotated candidate crops on 5,000 images varying in different topics (\eg lecture, gaming, VR, and vlog, \etc).
Massive experimental results on the \textit{UGCrop5K} dataset and other public benchmarks all reveal the superiority and effectiveness of our proposed network, which can outperform the state-of-the-art methods while keeping a good trade-off between speed and accuracy. Our contribution can be summarized as follows:

\begin{itemize}
\item 
We experimentally investigate the limitations of the existing cropping algorithms and analyze the main challenges in real-life applications. We then construct a new \textbf{\textit{UGCrop5K}} benchmark, to our best knowledge, which is the largest densely labeled cropping dataset with 450,000 high-quality annotated candidate crops. 
\item
We propose an efficient \textbf{\textit{S$^2$CNet}} with a modified adaptive attention graph to capture the relations between different objects in the images. By exploiting both semantic and spatial information, we can produce aesthetic cropped images and maintain content integrity.
\item 
Extensive experiments conducted on the proposed and other general datasets validate the merits of our approach against state-of-the-art cropping methods.
\end{itemize}

\section{Related Work}
\noindent \textbf{Aesthetic Image Cropping.}
Most of the early conventional works~\cite{suh2003automatic,stentiford2007attention,marchesotti2009framework,liu2010optimizing,zhang2013weakly,fang2014automatic} are based on hand-craft aesthetic features~\cite{li2006studying,ma2004automatic} and some criteria-based detection features such as face detection~\cite{zhang2005auto} and eye tracking~\cite{santella2006gaze}, \etc.
Later, benefiting from the deep learning models, more researchers pay attention to designing data-driven methods in various ways.
VFN~\cite{chen2017learning} proposed an end-to-end deep ranking net to implicitly model images. 
Later, Wei $\emph{et~al.}$~\cite{wei2018good} constructed a comparative photo composition (CPC) dataset for pairwise learning. 
However, pairwise learning cannot provide sufficient evaluation metrics for image cropping as pointed in~\cite{zeng2019reliable}. Recently, some works~\cite{wang2017deep,li2019image,tu2020image} exploited saliency detection~\cite{goferman2011context,hou2017deeply} to first locate the salient region and then generate candidate crops preserving the important content.
However, some of the complex UGC images and landscape photos have multiple salient objects or even no salient ones, which may lead to cropping failure~\cite{lu2019aesthetic}.
Hong $\emph{et~al.}$~\cite{hong2021composing} designed a dual branch network with the key composition map. However, it requires auxiliary composition datasets and manually defined rules, which lowers the upper bound of the usage. 
Zeng $\emph{et~al.}$~\cite{zeng2019reliable, zeng2020grid} introduced an efficient grid-based cropping method and proposed a densely annotated benchmark with new evaluation metrics.
Jia $\emph{et~al.}$~\cite{jia2022rethinking} formulated the task as object detection as DETR~\cite{carion2020end}. However, this kind of anchor generation approach can not yield specific aspect ratio crops. More recently, 
HCIC~\cite{zhang2022human} proposed a specific content-aware human-centric approach, which hinders its application for general object photos.

\begin{figure*}
 \begin{center}
  \centering
  \includegraphics[width=6.3in]{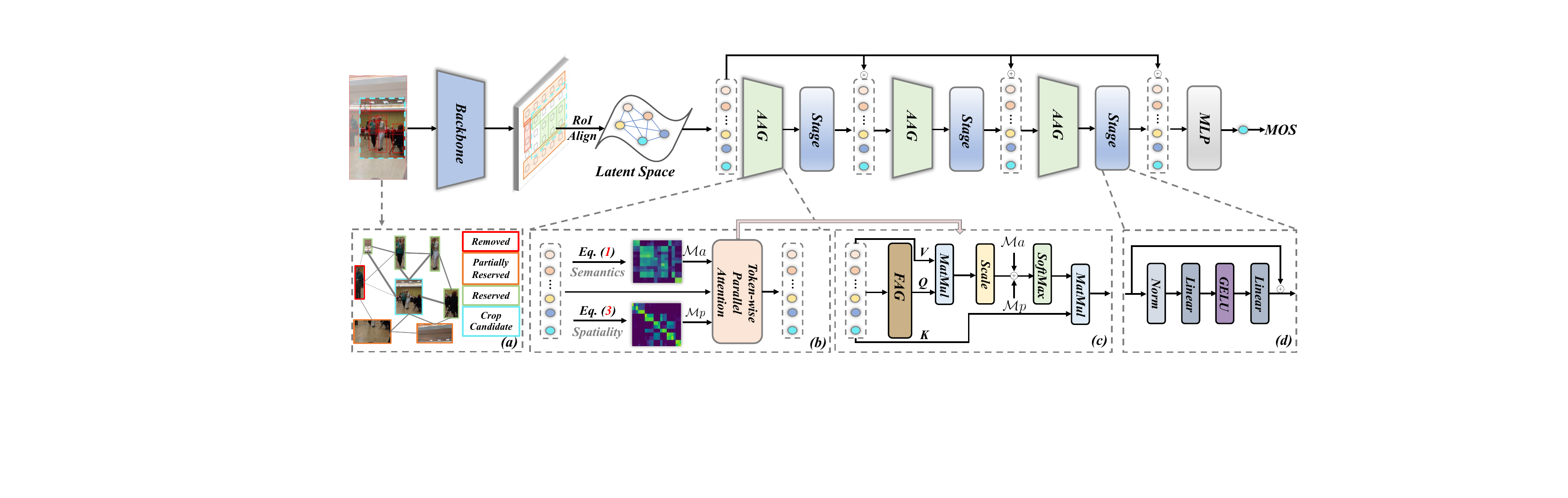}
 \end{center}
 \caption{The overall pipeline of our proposed framework. We first use the convolutional backbone to extract visual features followed by RoIAlign~\cite{he2017mask} and RoDAlign~\cite{zeng2019reliable} extracting $d$-dimension features for each potential object and the crop candidate. These features are then provided as inputs to the proposed adaptive attention graph (AAG), which performs joint spatial-semantic information propagation over each node in the graph. Ultimately, the updated messages are centralized to the crop candidate node to perform aesthetic score prediction.}\label{fig2}
\end{figure*}

\noindent \textbf{Region-based Relations Mining.}
Region-based relations mining is popular in visual tasks, which is widely used in video classification~\cite{wang2018videos}, segmentation~\cite{wang2019zero}, tracking~\cite{gao2019graph} and image outpainting~\cite{yang2022scene}, \etc. In the image cropping area, yet, rare works attempt to model the visual regional correlations. Although the most relevant approach CGS~\cite{li2020composing} proposed to model the mutual relations between the candidates, the global feature of each crop ignores the instance-wise information inside or outside the candidate, which fails to explicitly compose the visual elements and decides what should be preserved or abandoned. Besides, all the aforementioned strategies are usually built on graph~\cite{chen2020simple}, where they merely consider the semantic message of each node while neglecting the spatial location information. Furthermore, the conventional graph convolution networks will cause an over-smoothing problem when layers become deeper.
TransView~\cite{pan2021transview} later employed a vision transformer to capture image pixel-wise dependencies.
However, not all tokens are equally important in vision tasks as stated in~\cite{zeng2022not}. In addition, the self-attention mechanism in the transformer ignores the effective use of edge cues of nodes, such cues enjoy a good inductive bias~\cite{goyal2022inductive} for multi-object region learning.
To alleviate these issues, we propose a modified adaptive attention graph to perform image cropping, which can well model the instance-level relations and find aesthetically pleasing crops.

\section{Methodology}
\subsection{Network Overview}
Our motivation is based on explicitly building the compositional relations among the crop candidate and all object proposals. According to this, the network learns what should be \textit{removed}, \textit{partially reserved} and \textit{reserved} for an appealing crop that enjoys considerable content integrity.
As shown in Fig~\ref{fig2}(a), for the visual object outside the crop candidate (\eg the old man on the far left), since the person looks to the left, making the content semantically irrelevant to the crop and its aesthetic contribution is thereby weak. For the elements inside the crop candidate, we attempt to make the network capture mutual visually significant dependencies. And for some uncertain background objects, we learn to preserve the attractive parts while removing the redundant parts.
To achieve this goal, we adopt the adaptive attention graph (AAG) to model the scalable connections among the regional contents rather than using the plain transformer~\cite{dosovitskiy2020vit,pan2021transview} to model the visual patches equally.
 
Concretely, given an input image $\mathcal{I}$ corresponding with the crop candidate, we  leverage Faster RCNN~\cite{ren2015faster} pretrained on Visual Genome~\cite{krishna2017visual} to mine top-$N$ potential visual objects. We then obtain the feature map $\mathcal{F}$ by passing the image into the convolutional backbone~\cite{simonyan2014very,sandler2018mobilenetv2}. After that, we apply RoIAlign~\cite{he2017mask} and RoDAlign~\cite{gao2019graph} operations followed by the FC layer to get $d$-dimensional features of the total visual regions as $X = [x_1, x_2, ..., x_{N+1}] \in \mathbb{R} ^{(N+1) \times d}$ ($N$ detection boxes and one crop candidate). These features are then fed into our proposed network to capture high-order information. Finally, we predict the aesthetic score by aggregating the updated features.

\subsection{Adaptive Attention Graph}\label{321}
Formally, a graph $\mathcal{G}$ = ($\mathcal{V}$, $\mathcal{E}$) is defined as a set consisting of nodes and edges. Each node $v_i$ $\in$ $\mathcal{V}$ represents the extracted feature $x_i$, and each $e_{i,j}$ $\in$ $\mathcal{E}$ denotes the correlation between $v_{i}$ and $v_{j}$.
Note that we construct a fully connected graph since we consider that not only the relations between crop candidates and other regional objects are important, but the global relations among different visual objects also provide useful information for aesthetic composition.

\noindent \textbf{Semantic Edges.}
To represent the pair-wise relation among different nodes (\eg $x_i$ and $x_j$) and distribute different weights to the edges, we establish the relevance in an embedding space~\cite{vaswani2017attention} to compute the feature appearance similarity matrix $\mathcal{M}_a \in \mathbb{R}^{(N+1) \times (N+1)}$ as: 

\begin{equation}
\mathcal{M}_{a(i,j)} = \frac{\phi(x_i)^T \varphi(x_j)}{\sqrt{d}},\tag{1}
\label{eq1}
\end{equation}
where $\phi(x)$ = $W_{\phi}x + b_{\phi}$ and $\varphi(x)$ = $W_{\varphi}x + b_{\varphi}$ are two learnable linear functions that project the feature into the high-dimensional subspace. 

\noindent \textbf{Spatial Edges.} In addition to building the semantic relations among the nodes, the spatial information should also be considered since it contains useful topological representation. Specifically, we view the center coordinate $p_i$ = $(p_i^x, p_i^y)$ of the node $x_i$'s bounding box as an initial spatial feature. To this end, we explicitly model the spatial position connections of nodes in the following optional ways:

\textbf{\textit{DisDrop:}}  One assumption is that the nodes that are closer in space have more important information than nodes that are far away. And some of the distant nodes contribute less or even zero to aesthetic composition. 
In this way, we try to drop out the spatial relations of nodes whose distance exceeds a certain threshold and compute the spatial position matrix $\mathcal{M}_p \in \mathbb{R}^{(N+1) \times (N+1)}$ as follows:

\begin{equation}
\mathcal{M}_{p(i,j)} = \begin{cases}
\psi(\mathcal{D}(p_i, p_j))& {\textit{if} \ \ \mathcal{D}(p_i, p_j) \leq \epsilon * wid}\\
0& {\textit{if} \ \ \mathcal{D}(p_i, p_j) > \epsilon * wid}
\end{cases},\tag{2}
\label{eq2}
\end{equation}
where $\mathcal{D}(\cdot)$ is the \textit{Euclidean} distance calculating function. $\epsilon$ denotes the threshold, and $wid$ is the width of image. $\psi(\cdot)$ denotes a Multi-Layer Perceptron (MLP) layer that projects the $1$-dimensional distance to a high-dimensional vector.

\textbf{\textit{DisEmb:}} However, in some cases, the relations between the nodes that are farther away are also significant. For example, if two people are far away, but they are facing each other and have communication (\eg greeting or having eye contact), then the relation between these two nodes is much stronger than the closer nodes but without any connection.
Based on this observation, we formulate the spatial position matrix $\mathcal{M}_p$ as follows:

\begin{equation}
\mathcal{M}_{p(i,j)} = ||(W_m p_i + b_m) - (W_n p_j + b_n)||^2_2,\tag{3}
\label{eq3}
\end{equation}
where $W_{m;n}$ and $b_{m;n}$ are the different learnable weight matrices and biases that embed the distance into a vector.

\noindent \textbf{Correlation Adjacency.}
In order to jointly capture sufficient spatial-semantic information, we then construct a \textbf{spatial-semantic correlation} adjacency matrix $\mathcal{A} \in \mathbb{R}^{(N+1) \times (N+1)}$ combining both representations as follows:

\begin{equation}
\mathcal{A}_{(i,j)} = \frac{\mathcal{M}_{a(i,j)} \cdot {e} ^{\mathcal{M}_{p(i,j)}}}
{\sum_{j = 1}^{N+1} \mathcal{M}_{a(i,j)} \cdot {e} ^{\mathcal{M}_{p(i,j)}}}, \tag{4}
\label{eq4}
\end{equation}
where normalization is performed for each element. Thus, we have $\mathcal{A}_{i,j} \sim [0,1]$.

\subsection{Graph-Aware Attention Module}\label{322}
After assembling the graph, we perform the feature extraction over the nodes. 
As aforementioned, we modify the standard GCN to the graph-aware attention operation similar to Transformer~\cite{dosovitskiy2020vit} but merge the spatial-semantic features to generate the attention weights.

\begin{figure*}[t]
	\begin{center}
		\centering
		\includegraphics[width=6.3in]{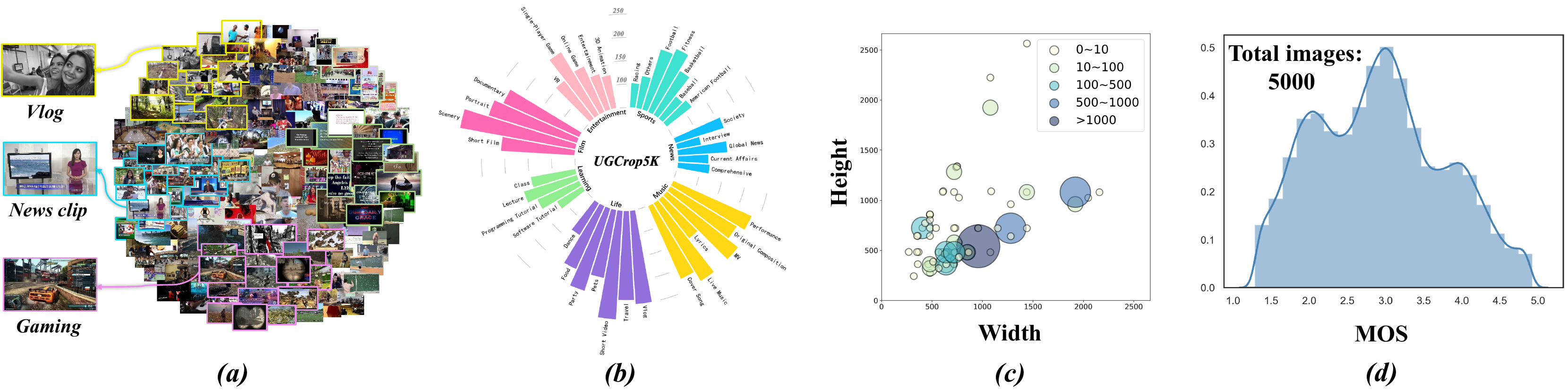}
	\end{center}
	\caption{Statistics of the proposed \textit{UGCrop5K} dataset, including (a) some visualization sample images, (b) taxonomic structure, (c) scatter plot of image width versus image height distribution with marker size indicating the number, and (d) histograms of the MOS.}
	\label{fig3}
\end{figure*}

\noindent \textbf{{Feature Aggregation Gate (FAG)}.} 
As depicted in Fig~\ref{fig2}(c), before calculating the self-attention of the node features, they are first fed into the feature aggregation gate to implicitly embed the information from the adjacency tensor. Specifically, we view the nodes as tokens. Considering the input feature $X$ and the correlation adjacency tensor $\mathcal{A}$, the scheme of FAG is computed as follows:

\begin{equation}
X = RELU(\mathcal{A}ZX), \tag{5}
\label{eq5}
\end{equation}
where $Z \in \mathbb{R}^{(N+1) \times d}$ is the learnable weight matrix. The output features $X$ aggregate the neighbouring node features, which can dynamically generate tokens with the appropriate importance to perform graph understanding.

\noindent \textbf{Spatial-Semantic Oriented Self-Attention (S$^2$O-SA).}
Afterwards, the outputs from FAG are viewed as queries $Q$, and the original nodes are used as keys $K$ and values $V$. We then reformulate the self-attention as follows:
\begin{equation}
S^2O \textit{-} SA = \textit{softmax}(\frac{QK^T}{\sqrt{d}} + \mathcal{M}_a + \mathcal{M}_p)V. \tag{6}
\label{eq6}
\end{equation}

By injecting both spatial and semantic edge features, it endows the self-attention mechanism with semantic-aware and topology-aware structures that models the nodes non-equally.
We omit the multi-head operation for clarity. In practice, we adopt several parallel multi-head attention to concatenate the features for better representation fusion.
Generally, the whole process can be stacked into multi-layers as follows:
\begin{equation}
\begin{split}
X &= FAG(X), \\
X' &= S^2O \textit{-} SA({LN}(X)) + X,  \\
X  &= {FFN}({LN}(X')) + X',
\end{split}
\tag{7}
\label{eq7}
\end{equation}
where $LN(\cdot)$ indicates the LayerNorm~\cite{ba2016layer} and $FFN$ is the feed-forward network.
Note that unlike the position encoding in the transformer that is added for sequential input data, the nodes in our paper are not arranged sequentially and are connected by edges.
The proposed spatial edge encodes the structural information in the self-attention of a graph with the capability to modulate the distance-related receptive field.

\subsection{Network Optimization}\label{323}
Ultimately, after obtaining the features from the adaptive attention graph, two layers of the MLPs are exploited to centralize the update message of all the nodes to the crop candidate to predict the aesthetic score. We first utilize the weighted smooth $\ell_1$ loss~\cite{ren2015faster} for score regression as follows:
\begin{equation}
\mathcal{L}_{pred} = \frac{1}{K} \sum_{i=1}^{K}  \ell_1 (y_i - \hat{y}_i), \tag{8}
\label{eq8}
\end{equation}
where $K$ is the number of the crop candidate within an image, $y_i$ and $\hat{y_i}$ are the predicted and ground-truth score of the $i$-th candidate view, respectively. 

In addition, following~\cite{li2020composing,zhang2022human}, we also use the ranking loss~\cite{chen2017learning} to explicitly learn the relative sorting orders between different crops as follows:
\begin{equation}
\mathcal{L}_{rank} = \frac{\sum_{i,j} \text{max}(0, \sigma(\hat{y}_i - \hat{y}_j)((y_i - y_j)(\hat{y}_i - \hat{y}_j))
)}{K(K-1)/2}, \tag{10}
\label{eq10}
\end{equation}
where $\sigma(\cdot)$ is the sign function. And the whole network is trained in an end-to-end manner.

\begin{table}[t]
\centering\fontsize{9pt}{10pt}\selectfont
        \begin{tabular}{cc|cc|cc}
        \toprule 
        \toprule 
        \multicolumn{2}{c|}{\multirow{2}*{\textit{Spatial Edge}}} &
        \multicolumn{2}{c|}{{\textbf{\textit{UGCrop5K}}}} &
        \multicolumn{2}{c}{{\textbf{\textit{GAIC}v1}}} \\
        \cline{3-6}
         & & $\overline{ACC_{5}}$ & $\overline{ACC_{10}}$ & $\overline{ACC_{5}}$ & $\overline{ACC_{10}}$ \\
        
        \midrule 
        \multirow{3}*{\textit{DisDrop} } & $\epsilon$ = 0.1 & 58.7 & 70.4 & 59.6 & 77.2 \\
          & $\epsilon$ = 0.2 &  60.4 & 71.6 & 60.5 & 77.8\\
         & $\epsilon$ = 0.3 &  59.6 & 70.7 & 59.9 & 77.4\\
        \rowcolor{mygray} {\textit{DisEmb}}& & \textbf{60.8} & \textbf{72.1} & \textbf{61.0} & \textbf{78.1}\\			
        \bottomrule 
    \end{tabular}
    \caption{Analysis of different spatial edges.}
    \label{table1}
\end{table}

\section{Experiment}
\subsection{Datasets and Metrics}
\textbf{\textit{Datasets}:} To fill the gap in the image cropping domain of real-life applications, we construct a large dataset, term as {\textbf{\textit{UGCrop5K}}}.
Specifically, we first collect parts of the user generated content from the opensource databases, including KoNViD-1k~\cite{hosu2017konstanz}, LIVE-VQC~\cite{sinno2018large,sinno2018large2,sinno2018large3}, YouTube-UGC~\cite{wang2019youtube} and Bilibili~\cite{livebot} social video websites. 
Furthermore, we also provide approximately 500 self-made contents using different devices (\eg shooting by iPhone 13 Pro Max, DJI Mavic 3 and Canon EOS R5 vertically or horizontally within various aspect ratios) from different scenarios to guarantee the variety of the proposed dataset. 
We then use HECATE~\cite{song2016click} to generate the top-$3$ cover images from the collected videos automatically.
For data cleaning, we manually remove low-quality ambiguous images (\ie pure colour images, highly similar content images, and blurry images) and reduce repetition. Particularly, we also remove images that are potentially not necessary for cropping.
Finally, we have a total of 5,000 images with different aspect ratios covering different scenes, as shown in Fig~\ref{fig3}(a) $\sim$ (c).

Subsequently, 20 annotators are invited to our image composition annotation task, including 3 non-professional students, 10 medium-professional people engaged in art-related studies and 7 experienced workers in photography.
We generate 90 predefined anchor boxes similar to~\cite{zeng2020grid} for each image and develop an online website annotation tool instead of the annotation software~\cite{zeng2020grid} that depends on the specific computer environment to ease the burden of the annotators. 
Concretely, annotators assign each predefined crop an integer score ranging from 1 to 5, with higher scores representing the better composition. Each crop needs to be rated by at least 5 people. We finally calculate the mean opinion score (MOS) for each candidate crop as its ground-truth quality score, and Fig~\ref{fig3}(d) shows the histograms of the MOS.
In general, we have 5,000 images with 450,000 high-quality annotated candidate crops in the dataset, and we split 4,200 images for training and 800 images for testing. Due to resource constraints, we conducted experiments on an early version of the dataset annotated by partial annotators. The complete dataset and results will be released in Github. To verify the generalization of our model, we also conduct experiments on other public image cropping benchmarks: {\textbf{\textit{GAIC}v1}}~\cite{zeng2019reliable} and {\textbf{\textit{GAIC}v2}}~\cite{zeng2020grid} datasets.

\begin{table}[t]
\centering\fontsize{9pt}{10pt}\selectfont
        \begin{tabular}{c|cc|cc}
        \toprule 
        \toprule 
        \multirow{2}*{\textit{Object Proposal}}  &
        \multicolumn{2}{c|}{{\textbf{\textit{UGCrop5K}}}} &
        \multicolumn{2}{c}{{\textbf{\textit{GAIC}v1}}} \\
        \cline{2-5}
        & $\overline{ACC_{5}}$ & $\overline{ACC_{10}}$ & $\overline{ACC_{5}}$ & $\overline{ACC_{10}}$ \\
        
        \midrule 
        {\textit{N} = 8}  &  59.9 & 71.2 & \textbf{61.2} & 78.0\\
        \rowcolor{mygray} {\textit{N} = 10} & \textbf{60.8} & \textbf{72.1} & 61.0 & \textbf{78.1} \\
        {\textit{N} = 12} & 60.6 & 71.9 & 59.6 & 77.4 \\
        {\textit{N} = 15} & 60.4 & 71.5 & 58.7 & 77.0\\	
        \bottomrule
    \end{tabular}
    \caption{Analysis of the object proposal number.}
        \label{table3}
\end{table}

\vspace{1.0ex}

\noindent \textbf{\textit{Metrics}:} Following~\cite{zeng2019reliable,zeng2020grid}, we adopt the averaged Spearman’s Rank-order Correlation Coefficient ({$\overline{SRCC}$}) and the averaged top-$k$ accuracy ($\overline{ACC_{k}}$) for both $k$ = 5 and $k$ = 10 as evaluation metrics instead of the unreliable Intersection-over-Union ({IoU}) metric.

\vspace{1.0ex}

\subsection{Implementation Details}
Following the existing methods~\cite{zeng2020grid,pan2021transview,zhang2022human}, we adopt MobileNetV2~\cite{sandler2018mobilenetv2} pretrained on ImageNet~\cite{deng2009imagenet} as backbone to extract multi-scale feature map.
The short side of the input sample is resized to 256 and maintains the aspect ratio. The aligned size of RoIAlign is set to $15 \times 15$ and the proposal number is set to 10 empirically. We stack 2 layers of the adaptive attention graphs with the multi-head number of 4.
The network is optimized by AdamW with the learning rate of 1e-4 for 80 epochs.
Data augmentations are similar to prior works~\cite{zeng2020grid,li2020composing}, including random flipping, saturation, and lighting noise are adopted.

\subsection{Ablation Analysis}

\begin{table}[t]
    \centering\fontsize{9pt}{10pt}\selectfont
            \begin{tabular}{ccc|cc|cc}
            \toprule 
            \toprule 
            \multirow{2}*{$\mathcal{G}$}&\multirow{2}*{$\mathcal{M}_{a}$}&\multirow{2}*{$\mathcal{M}_{p}$}&
            \multicolumn{2}{c|}{{\textbf{\textit{UGCrop5K}}}} &
            \multicolumn{2}{c}{{\textbf{\textit{GAIC}v1}}} \\
            \cline{4-7}
            & & & $\overline{ACC_{5}}$ & $\overline{ACC_{10}}$ & $\overline{ACC_{5}}$ & $\overline{ACC_{10}}$ \\
            
            \midrule
              &  &  & 54.3 & 64.8 & 49.7 & 68.4 \\
            \ding{52} &  &  & 55.2 & 65.7 & 51.1 & 70.5\\
             & \ding{52} & & 54.7 & 65.3 & 50.8 & 70.2\\
             &  & \ding{52} & 54.9 & 65.8 & 53.4 & 72.6\\
            \ding{52} & \ding{52} &  & 58.7 & 69.6 & 56.7 & 76.1\\
            \ding{52} &  & \ding{52} & 59.2 & 70.9 & 58.9 & 77.0\\
             & \ding{52} & \ding{52} & 60.0 & 71.5 & 60.2 & 77.4\\
            \rowcolor{mygray}\ding{52} & \ding{52} & \ding{52} & \textbf{60.8} & \textbf{72.1} & \textbf{61.0} & \textbf{78.1} \\
            \bottomrule 
        \end{tabular}
        \caption{Analysis of different proposed components. $\mathcal{G}$ indicates FAG module.}
        \label{table2}
\end{table}

\begin{table}[t]
    \centering\fontsize{9pt}{10pt}\selectfont
        \begin{tabular}{c|ccc}
        \toprule 
        \toprule 
        \multirow{2}*{\textit{Ratios}}  &
        \multicolumn{3}{c}{{\textbf{\textit{Perc. (\%)}}}} \\
        \cline{2-4}
        & \textit{Ours} & Mars &GAICv2  \\
        \midrule 
        {{3:4}} & \cellcolor{mygray}\textbf{{47.7}} & 35.3 & 17.0  \\
        {{4:3}} &  \cellcolor{mygray}\textbf{{44.0}} & 24.3 & 31.7  \\
        {{16:9}} &  \cellcolor{mygray}\textbf{{52.0}} & 23.3 & 24.7 \\
        \bottomrule
        \end{tabular}
        \caption{User study results.}
        \label{table5}
\end{table}

\begin{figure}[t]
    \centering
    \includegraphics[width=3.0in]{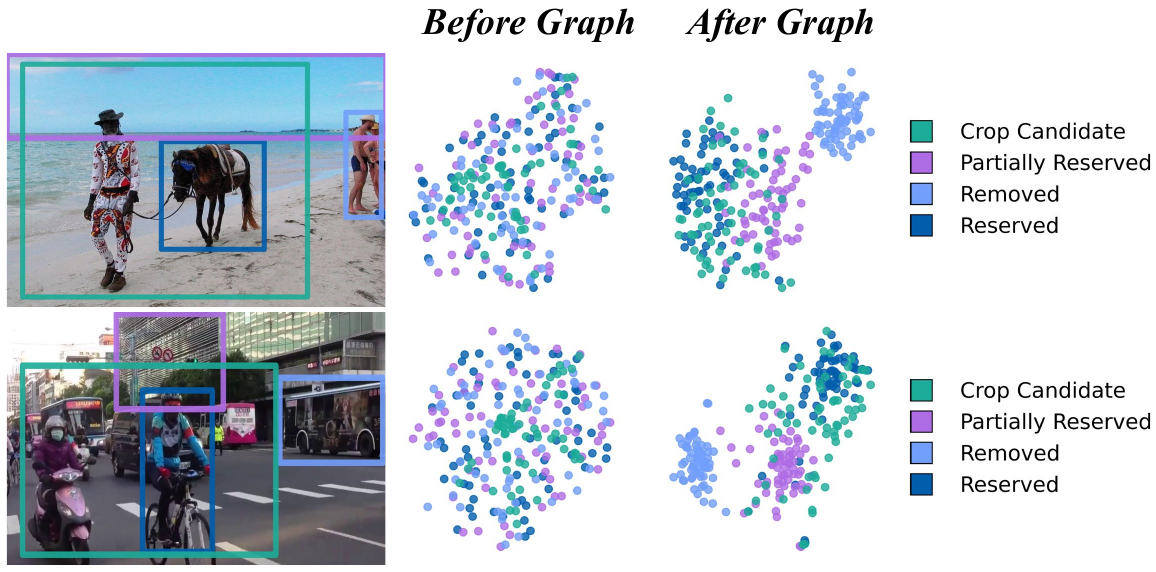}
    \caption{The t-SNE feature visualization before and  after the proposed graph. Different colours indicate the crop candidates, the regions should be removed, reserved, or partially reserved, respectively. The features before the graph show indistinguishable clusters, while the features learned by our graph are more discriminative, which can guide the model to find good views more reasonably. Zoom in for the best view.}
    \label{fig4}
\end{figure}

\vspace{1.0ex}

\noindent \textbf{Exploration of different spatial edges:}
As shown in Table~\ref{table1}, we first explore different constructions of spatial edge mentioned before. It shows that when $\epsilon$ = 0.2, $\textit{DisDrop}$ can yield relatively good performance. Although different thresholds $\epsilon$ can be altered, $\textit{DisEmd}$ can always outperform $\textit{DisDrop}$. 
As aforementioned, the reason for the above observation is that the mutual contributions of  visual features at different locations weakened by distance may miss some high-order spatial information. Therefore, in the following paper, we adopt the $\textit{DisEmd}$ strategy to build the spatial edge in a more holistic manner. 

\vspace{1.0ex}

\begin{table*}[t]
	\begin{center}\fontsize{9pt}{10pt}\selectfont
			\begin{tabular}{ccc|ccc|ccc|ccc}
			\toprule 
			\toprule 
			\multirow{2}*{\textit{Model}} \
		    &\multirow{2}*{\textit{Param (M)}} \ & \multirow{2}*{\textit{FPS} $\uparrow$}  &
			\multicolumn{3}{c|}{{\textbf{\textit{UGCrop5K}}}} &
			\multicolumn{3}{c|}{{\textbf{\textit{GAIC}v1}}} &
                \multicolumn{3}{c}{{\textbf{\textit{GAIC}v2}}} \\
			\cline{4-12}
			& & & $\overline{SRCC}$ & $\overline{ACC_{5}}$ & $\overline{ACC_{10}}$ & $\overline{SRCC}$ & $\overline{ACC_{5}}$ & $\overline{ACC_{10}}$ & $\overline{SRCC}$ & $\overline{ACC_{5}}$ & $\overline{ACC_{10}}$ \\
			
			\midrule
                VFN & 11.55 & 0.4 & 0.372 & 25.4 & 36.1 & 0.450 & 26.7 & 38.7 & 0.485 & 26.4 & 40.1\\
			A2-RL & 24.11 & 2.6 & - & 22.8 & 33.9 & - & 23.0 & 38.5 & - & 23.2 & 39.5\\
                VEN & 40.93 & 0.3 & 0.394 & 31.3 & 42.7& 0.621 & 37.6 & 50.9 & 0.616 & 35.5 & 48.6\\
                VPN & 65.31 & 96.2 & - & 35.8 & 44.3 & - & 40.0 & 49.5 & - & 36.0 & 48.5\\
                GAICv1 & 13.54 & 129.8 & 0.418 & 45.7 & 52.8 & 0.735 & 46.6 & 65.5 & 0.832 & 63.5 & 79.0\\
                GAICv2 & \textbf{1.81} &  \textbf{212.4} & 0.466 & 54.7 & 64.5 & 0.783 & 57.2 & 75.5 & 0.849 & 63.9 & 79.7\\
                ASM-Net & 14.95 & 102.0 & 0.435$^\dag$ & 52.8$^\dag$ & 63.2$^\dag$ & 0.766 & 54.3 & 71.5 & 0.837$^\dag$ & 63.2$^\dag$ & 79.1$^\dag$\\
                CGS & 13.68 & 100.0  & 0.467 & 56.4 & 66.8 & \textbf{0.795} & \underline{59.7} & \underline{77.8} & 0.848 & 63.5 & 79.4\\
                TransView & 4.62 & 147.3 & \underline{0.482}$^\dag$ & \underline{57.9}$^\dag$ & \underline{69.4}$^\dag$ & 0.789$^\dag$ & 59.2$^\dag$ & 77.4$^\dag$ & 0.857 & \underline{63.9} & 82.4\\
                HCIC & 19.47 & 128 & 0.449 & 54.5 & 64.1 & \underline{0.793} & 58.6 & 74.5 & 0.851 & 63.8 & 81.3\\
                SFRC & 5.91 & 40 & - & - & - & - & - & - & \textbf{0.865} & 63.7 & \underline{82.6}\\
			\midrule
                \rowcolor{mygray} \textbf{\textit{S$^2$CNet}} & \underline{3.92} & \underline{162.8} & \textbf{0.502} & \textbf{60.8} & \textbf{72.1} & \underline{0.793} & \textbf{61.0} & \textbf{78.1} & \underline{0.861} & \textbf{64.0} & \textbf{82.7}\\
                \bottomrule     
	    \end{tabular}
     \caption{Quantitative comparison to other state-of-the-art approaches on \textit{UGCrop5K}, \textit{GAIC}v1 and \textit{GAIC}v2 datasets. The best performance is in bold, and the second-best is underlined. $\dag$ indicates our re-implement results since the authors do not provide codes. Other results are derived from the open-source codes and the original papers.}
     \label{table4}
	\end{center}
\end{table*}

\begin{figure*}[t]
	\begin{center}
		\centering
		\includegraphics[width=7in]{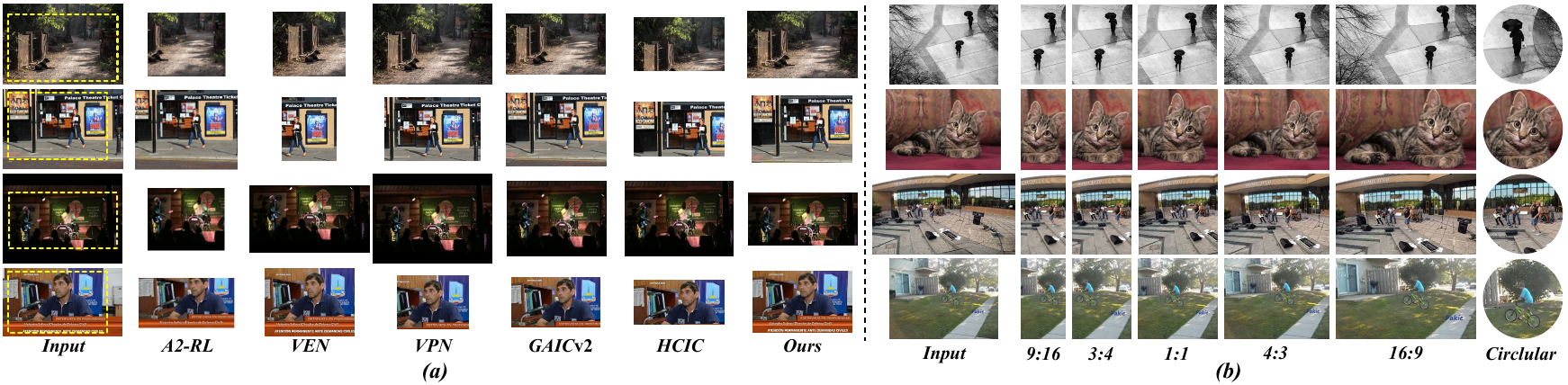}
	\end{center}
	\caption{(a): Qualitative comparisons of different state-of-the-art methods. The first two rows of images are from the \textit{GAIC}v1 and \textit{GAIC}v2 datasets, and the last two rows of images are from the \textit{UGCrop5k} dataset. The top-scored best crops are in the yellow dotted box. (b): Image cropping results with different aspect ratios.}
	\label{fig5}
\end{figure*}

\noindent \textbf{Exploration of object proposal number.} We also explore the effect of
 different object proposal numbers, as shown in Table~\ref{table3}. When $N$ = 10, we can achieve satisfactory results on both datasets. When $N$ = 8, $\overline{ACC_5}$ can be improved in \textit{GAIC}v1 benchmark since it does not contain many objects as in \textit{UGCrop5k}. 
 When continuously increasing the $N$, the cropping performance will drop. We conclude that too many object proposal features are redundant, which may confuse the network for learning effective relations.

\vspace{1.0ex}

\noindent \textbf{Exploration of different proposed components.}
Table~\ref{table2} analyzes that each component can boost our network to varying degrees compared to the \textit{baseline} (our \textit{baseline} depends on pure Transformer block). Particularly, + $\mathcal{M}_{p}$ obtains more improvement than  + $\mathcal{M}_{a}$, we conjecture that the topological spatial cues are more unique and useful in the self-attention operation.
By combining all the components, we can achieve the best performance, which verifies the proposed modules are helpful and indispensable.

\vspace{1.0ex}

\noindent \textbf{Exploration of different graphs.} we further compare our proposed graph with the conventional GCN~\cite{li2020composing} and GAT~\cite{brody2022how}. After replacing our graph with GCN, $\overline{ACC_5}$ will drop from
$\textbf{60.8} \xrightarrow[]{-4.6} \textbf{56.2}$ and $\textbf{61.0} \xrightarrow[]{-3.7} \textbf{57.3}$ on \textit{UGCrop5k} and \textit{GAIC}v1 benchmarks, respectively. When replacing with GAT, $\overline{ACC_5}$ will drop to $\textbf{59.4}$ and $\textbf{58.6}$ on two above benchmarks.
This validates the effectiveness of our proposed adaptive attention graph.

\vspace{1.0ex}

\noindent \textbf{Model interpretability:} As shown in Fig~\ref{fig4}, we show how our proposed adaptive attention graph encodes the information. The top-$1$ crop candidate and some detected object proposals are depicted in the leftmost image. 
By comparing the feature distribution maps~\cite{van2008visualizing}, we can observe that the different regional features will diffuse or aggregate rather than in a mixed cluster. More specifically, the crop view features are closer to reserved features and have a certain degree of overlap with the partially reserved ones while far away from the removed features. 
This is because the relations between the crop view and aesthetically unnecessary contents are weakened via graph learning, and the edge weights between the crop view and potentially necessary contents are strengthened. 
This can help the network explicitly discriminate good and bad views through backpropagation learning with annotation scores.

\subsection{Compare with the State-of-the-art Methods}
\noindent \textbf{Quantitative Results.}
As shown in Table~\ref{table4}, \textbf{\textit{S$^2$CNet}} can not only outperform state-of-the-art methods on the proposed challenging \textit{UGCrop5k} dataset but also achieve satisfactory results on two other general \textit{GAIC}v1 and \textit{GAIC}v2 benchmarks. 
Nevertheless, our network can also achieve reliable results in general scenarios, which demonstrates the effectiveness and soundness of the proposed network. 
Note that SFRC~\cite{wang2023image} utilized additional unlabeled test data for training. For fair comparisons, we report its performance under the inductive setting.
Besides, we also report the model complexity and runtime, all the experiments are executed on a single NVIDIA RTX 2080Ti GPU. Our method can process images at rates of 162.8 FPS while keeping competitive results, which guarantees the efficiency and practicality of the network.

\noindent \textbf{Qualitative Analysis.} Fig~\ref{fig5}(a) shows the qualitative comparisons, from which we can observe that: (\textbf{i}) Our method can produce more aesthetically pleasing cropped views. They not only retain the main foreground of the photos but also can effectively preserve or remove some areas of the background to a greater extent for composition, and it is closer to the best annotated ground-truth; (\textbf{ii}) Our method can maintain image content integrity. As shown in the last row in Fig~\ref{fig5}(a), although other methods successfully crop the main person and achieve a relatively good view, they lose some useful attributes of the image (\ie A2-RL~\cite{li2018a2}, VEN/VPN~\cite{wei2018good} and GAICv2~\cite{zeng2020grid} cut out the important theme text of the news; HCIC~\cite{zhang2022human} even only keeps the main person), which may deliver incomplete information to readers.

\noindent \textbf{Applications.}
In real-life applications, cropping is usually constrained. We then visualize our cropping results with different common aspect ratios and one more circular view.
For circular cropping, we do not need to re-train our network, and we use the circular circumscribed square as the crop candidate to pass through our network.
As shown in Fig~\ref{fig5}(b), our model can find good views under different constraints, which demonstrates the ability of our model and meets the demand for UGC cropping, including cover image cropping, thumbnailing and icon generation, \etc.

\vspace{1.0ex}
\noindent \textbf{User Study.}
To evaluate the qualities of views within specific aspect ratios, we compare the proposed method
with other approaches (\eg Mars~\cite{li2020learning} and GAICv2~\cite{zeng2020grid}) that can also handle specific ratio cropping through the subjective user study.
We randomly collect 100 images and 200 images from \textit{GAIC}v2~\cite{zeng2020grid} and \textit{UGCrop5K} datasets. Then 15 volunteers are invited to select their favourite crop view from the results. Note that the experts are unaware of the views produced from which algorithms for fair comparisons. Table~\ref{table5} shows that our method can achieve the highest percentage and outperform the other methods.

\section{Conclusion}
In this paper, we introduce a spatial-semantic collaborative cropping network for user generated content and conduct a large densely labeled \textit{UGCrop5k} dataset for follow-up research in the cropping domain.
By exploring the semantic appearance and spatial topology information of different visual patches, we address the cropping task from a comprehensive perspective.
Extensive experiments on different datasets show that our method outperforms the existing cropping approaches qualitatively and quantitatively. 

\section{Acknowledgments}
This work was supported by National Natural Science Foundation of China (NSFC) 62272172, Guangdong Basic and Applied Basic Research Foundation 2023A1515012920. This work is supported in part by a Tencent Research Grant and National Natural Science Foundation of China (No. 62176002).

\bibliography{aaai24}

\end{document}